%% file: sample-sigconf.tex
\documentclass[sigconf]{acmart}
\hypersetup{bookmarksdepth=-2}
\usepackage{booktabs} 
\usepackage{listings}
\usepackage{subcaption}
\usepackage{multirow}
\usepackage{soul}

\usepackage{balance}
\usepackage[shortlabels]{enumitem}

\copyrightyear{2018}
\acmYear{2018}
\setcopyright{acmlicensed}
\acmConference[ICSE-NIER'18]{40th International Conference on Software Engineering: New Ideas and Emerging Results Track}{May 27-June 3 2018}{Gothenburg, Sweden}
\acmBooktitle{ICSE-NIER'18: 40th International Conference on Software Engineering: New Ideas and Emerging Results Track, May 27-June 3 2018, Gothenburg, Sweden}
\acmPrice{15.00}
\acmDOI{10.1145/3183399.3183427}
\acmISBN{978-1-4503-5662-6/18/05}

\usepackage{color}

\definecolor{dkgreen}{rgb}{0,0.6,0}
\definecolor{gray}{rgb}{0.5,0.5,0.5}
\definecolor{mauve}{rgb}{0.58,0,0.82}

\makeatletter
\let\origsection\section
\renewcommand\section{\@ifstar{\starsection}{\nostarsection}}

\newcommand\nostarsection[1]
{\sectionprelude\origsection{#1}\sectionpostlude}

\newcommand\starsection[1]
{\sectionprelude\origsection*{#1}\sectionpostlude}

\newcommand\sectionprelude{%
  \vspace{-1ex}
}

\newcommand\sectionpostlude{%
  \vspace{-1ex}
}
\makeatother

\newcommand{\inlinedComment}[2]
	{\textcolor{#1}{\small\textbf{#2}}}

\newcommand{\lx}[1]{\inlinedComment{red}{JLX says: #1}}

\begin{document}
\title{Hierarchical Learning of Cross-Language Mappings \\ through Distributed Vector Representations for Code}

\author{Nghi D. Q. Bui}

\affiliation{%
  \department{School of Information Systems}
  \institution{Singapore Management University}
}
\email{dqnbui.2016@phdis.smu.edu.sg}

\author{Lingxiao Jiang}
\affiliation{%
  \department{School of Information Systems}
  \institution{Singapore Management University}
}
\email{lxjiang@smu.edu.sg}


\begin{abstract}
\input{abstract}
\end{abstract}

%
%

\maketitle

\vspace*{-10pt}
\input{samplebody-conf}

\let\oldthebibliography=\thebibliography
\let\endoldthebibliography=\endthebibliography
\renewenvironment{thebibliography}[1]{%
  \begin{oldthebibliography}{#1}%
    \fontsize{7.2}{7.8}\selectfont
    \setlength{\parskip}{0ex}%
    \setlength{\itemsep}{0pt}%
}%
{%
  \end{oldthebibliography}%
}

\balance
\bibliographystyle{ACM-Reference-Format}
\bibliography{sample-bibliography}

\end{document}

%% file: abstract.tex
Translating a program written in one programming language to another can be useful for software development tasks that need functionality implementations in different languages.
Although past studies have considered this problem,
they may be either specific to the language grammars, or specific to certain kinds of code elements (e.g., tokens, phrases, API uses).
This paper proposes a new approach
to automatically learn cross-language representations for various kinds of structural code elements that may be used for program translation.
Our key idea is two folded:
First, we normalize and enrich code token streams with additional structural and semantic information,
and train cross-language vector representations for the tokens 
(a.k.a.\ {\em shared embeddings} based on word2vec, a neural-network-based technique for producing word embeddings;
Second, hierarchically from bottom up, we construct shared embeddings for code elements of higher levels of granularity (e.g., expressions, statements, methods) from the embeddings for their constituents, and then build mappings among code elements across languages 
based on similarities among embeddings.

Our preliminary evaluations on about 40,000 Java and C\# source files from 9 software projects show that our approach can automatically learn shared embeddings for various code elements in different languages and identify their cross-language mappings with reasonable Mean Average Precision scores.
When compared with an existing tool for mapping library API methods, our approach identifies many more mappings accurately. The mapping results and code can be accessed at {\scriptsize \url{https://github.com/bdqnghi/hierarchical-programming-language-mapping}}.
We believe that our idea for learning cross-language vector representations with code structural information can be a useful step towards automated program translation.

%% file: samplebody-conf.tex
\section{Introduction}
\label{sec:intro}
\input{intro}

\section{Related Work}
\label{sec:related}
\input{related}

\section{Our Approach}
\label{sec:approach}
\input{approach}

\section{Empirical Evaluation}
\label{sec:expr}
\input{expr}

\input{discussion}

\section{Conclusion and Future Work}
\label{sec:conclusion}
\input{conclusion}

\section{Acknowledgements}
\label{sec:ack}
\input{ack}

%

%% file: intro.tex
Automated program translation (a.k.a. language migration) can be very useful for software development as it may help reduce developer coding time, especially for functionalities and library APIs that need to be implemented and maintained in various programming languages.
Take the Apache Lucene as an example: it is a popular information retrieval library, providing many APIs for a third-party client program to access its core functionalities. Lucene was originally implemented in Java and not easy to be used by client programs in other languages. Due to popular demands for its functionalities, it has been ported to other languages (e.g., C\#, C++, Python, Ruby, PHP, etc.) to support clients in those languages. Nevertheless, multiple versions of Lucene in different languages increase the cost on its maintenance and development, as new features or bug fixes in one version may need to be manually ported to another version for consistency.
An automated approach to translate code among languages can help save much cost, and still be useful even if the approach cannot generate complete translations but can identify likely translation candidates in large code bases.

Existing studies on program translation may be classified in two categories. One is based on grammar rules (e.g., {\scriptsize Java2CSharp at https://github.com/codejuicer/java2csharp}), which can be very accurate, but inflexible in dealing with different languages or language evolutions as the translations need to be programmed repeatedly for different grammars.
The other is based on statistical language models for selected code elements (e.g., for tokens~\cite{Nguyen2013}, token phrases with contexts~\cite{Nguyen2015,Nguyen2016}, or APIs and API sequences~\cite{Zhong2010,Zhong2013,Nguyen2014,DBLP:conf/kbse/NguyenNNN14,Phan2017,Gu2016,Gu2017}), which can deal with different languages but may need to incorporate various kinds of contexts (e.g., sequences/co-occurrence relations, data/control dependencies) for selected code elements~\cite{Nguyen2016,Gu2017}.

In the field of natural language processing (NLP), neural-network-based Neural Machine Translation (NMT) has emerged as an alternative to statistical language models, achieving good results for natural language translation~\cite{DBLP:conf/nips/SutskeverVL14}.
NMT models use distributed vector representations of words as the basic unit to compose representations for more complex language elements, such as sentences and paragraphs.
One prominent distributed vector representation is word2vec \cite{Mikolov2013word2vec,mikolov2013distributed}, which uses neural networks to learn vector representations of words (a.k.a.\ {\em word embeddings}) from natural language articles to capture latent semantics with respect to a modeling objective, such as predicting the context given a word or predicting the next word given a context.
Also, similarities among different natural languages can be exploited for machine translation~\cite{mikolov2013exploiting}, which can be applicable for programming languages as there are many across-language code clones too~\cite{Cheng2016}.

Inspired by NMT and limitations in existing program translation techniques, {\bf our goal} is to produce a new way of {\em representing code in distributed vectors for any kind of code elements across languages}.

{\bf Our key idea} is mainly based on two observations: (1) code clearly has hierarchical structures as illustrated in Figure~\ref{figure:softwarelayers} and is often composable, and (2) code structures (in addition to its textual appearance) often accurately reflect its semantics, which are different from natural languages. That means, NMT may be able to generate distributed vector representations that can closely reflect the code semantics if the code token streams can be enriched with its structural information, and generate vector representations for any composed code elements that are of higher levels of granularity.
Therefore, {\bf our approach} works by normalizing and enriching code token streams with structural (and some semantic) information extracted via code parsing, constructing a bilingual skip-gram model to generate distributed vectors for code tokens in two different languages (a.k.a.\ {\em shared embeddings}), and composing shared embeddings for low-level code elements into more complex ones according to code structures.
Code elements in different languages but having similar shared embeddings will thus become mapping and translation candidates for each other.

{\bf Our preliminary evaluations} using about 40,000 source files from 9 programs that have multiple versions in both Java and C\# show that our approach can automatically learn shared embeddings from existing code across Java and C\#, and achieve around 50\% precision in recommending top-10 cross-language code mappings at various levels of granularity.
Compared with existing tools for identifying API mappings (StaMiner~\cite{DBLP:conf/kbse/NguyenNNN14}), our approach can identify more than 400 more library API methods and classes accurately.

\textbf{{Our main contributions}} in this paper are as follows:
\begin{itemize}[leftmargin=1em,topsep=0pt,itemsep=0pt,parsep=0pt,partopsep=0pt]
	\item We propose a new way to add structural information into source code token streams and adapt word embeddings to learn vector representations for code tokens across languages.
    \item We allow hierarchial compositions of vector representations for simpler code elements into more complex ones according to code structures, and thus can produce vector representations for any code structures across languages.
\end{itemize}


%% file: related.tex
There have been many studies on code representation, modeling, and learning for various purposes.
\citet{hellendoorn2017deep} point out that simpler models (e.g., n-gram) with caches of code locality and hierarchy may even outperform complex models (e.g., deep neural networks). But this may also indicate that using code locality and structural information within deep learning may further improve code learning accuracy. \citet{DBLP:conf/kbse/NguyenNNN14} also demonstrates that semantic code features at various levels
of abstraction can be useful in improve the accuracy of statistical language models. All of these studies provoke us to use structural information too when learning vector representations for code.

More specifically for program translation, among much work that utilizes statistical models for various kinds of code elements (e.g., tokens and token sequences~\cite{Nguyen2013,Nguyen2015,Nguyen2016}, and APIs~\cite{Zhong2013,Nguyen2014,DBLP:conf/kbse/NguyenNNN14,Phan2017,Zhong2010}), only a few have used deep learning~\cite{Gu2017,Gu2016}, but they are limited to API sequences, and may require much adaptation for different kinds code structures, while our approach may be applicable to learn shared embeddings for any kind of composable code structures.

%% file: approach.tex
\begin{figure}[t!]
    \centering
	\includegraphics[width=0.40\textwidth]{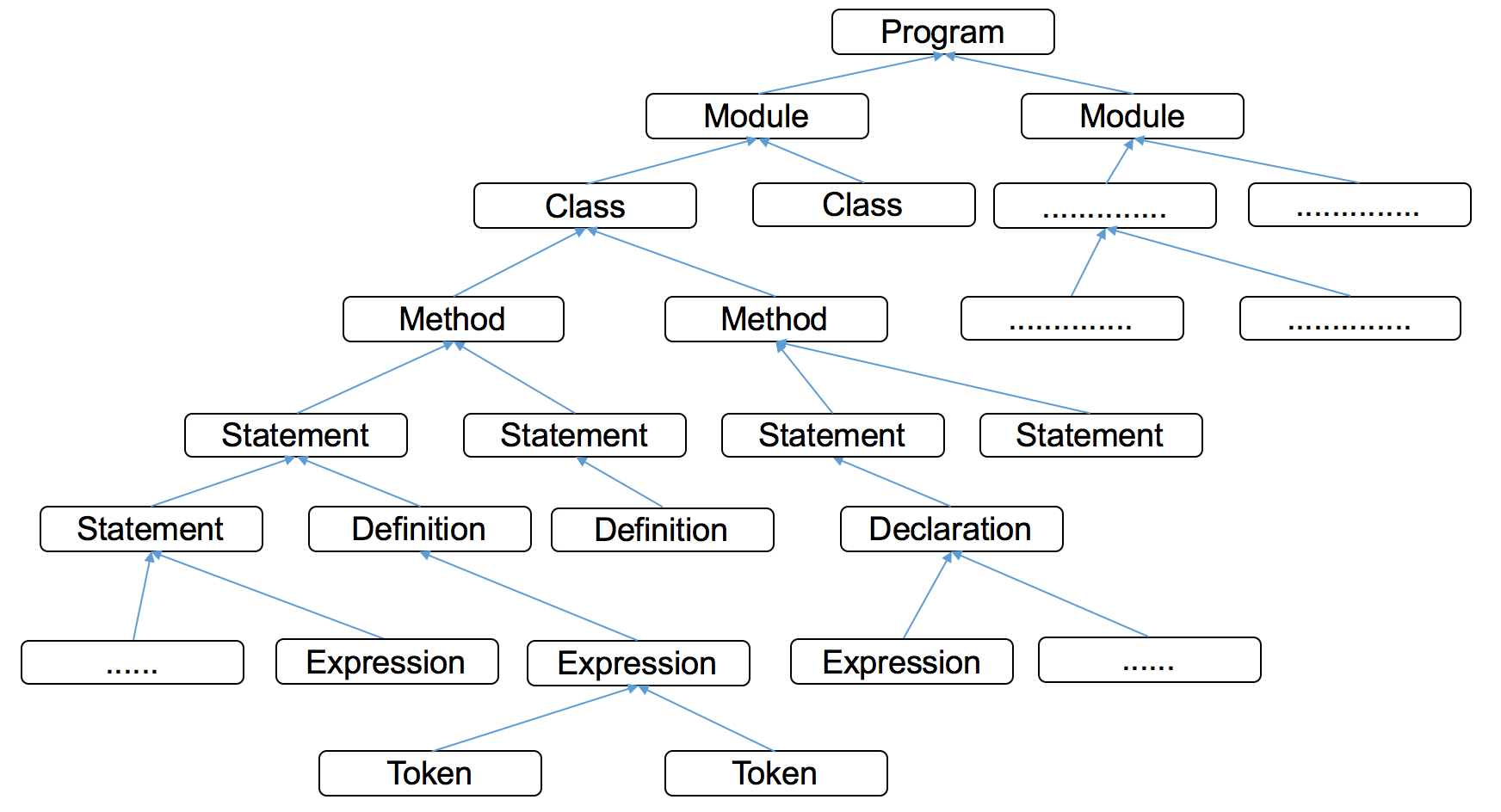}
	
	\caption{Abstract syntax trees --- illustration of code hierarchical structure and composability.}
	\label{figure:softwarelayers}
\end{figure}

\subsection{Overview}
\begin{figure*}[t!]
	\includegraphics[width=0.80\textwidth]{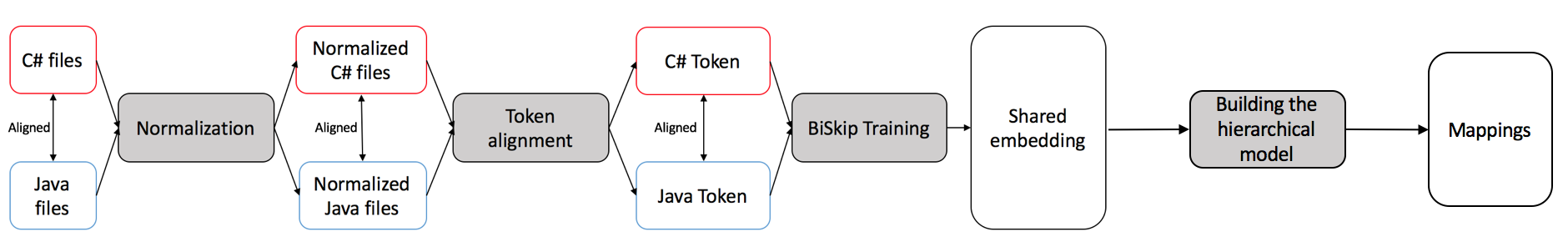}
    \vspace{-7pt}
	\caption{An overview of our approach}
	\label{fig:approach}
\end{figure*}

Figure \ref{fig:approach} is the overview of our approach. We first collect parallel corpus across languages for training bilingual embedding models. A parallel corpus is a collection of source code in one language and their translation into another language.
We utilize the similarity among file names to identify files in different languages that implement a same functionality.
Taking Lucene as an example, the file {\small \texttt{AbstractEncoder.cs}} in its C\# version has the same name as the file {\small \texttt{AbstractEncoder.java}} in its Java version. Thus, the files in the parallel corpus are considered to be ``semantically aligned'' with each other and used for later steps.
We then normalize the token streams in the files to remove semantic-irrelevant information (e.g., some variable names) and add more structural and  some semantic information (e.g., syntactic node types, data types, and method signatures).
The normalized token streams are then used as input for the Berkeley aligner \cite{liang2006alignment} to generate token-level alignment information indicating potential synonyms across languages,
and the token-aligned data is then used to learn bilingual vector representations for the tokens.
Finally, vector representations for low-level tokens are composed together to form representations for code elements of higher levels of granularity.
Code elements of similar vector representations across languages (i.e., shared embeddings) will be identified as mapping candidates for each other.

\vspace*{-1ex}
\subsection{Token Normalization}
\label{step:normalization}
This step (1) converts each raw token into its signature version and (2) adds structural keywords for the tokens based on ASTs.

\textbf{Convert a raw token into its signature:} This is to normalize the effects of various kinds of identifier names as some names are important for code semantics while some others are not.
For example, \texttt{class Text} in Lucene and \texttt{class Text} in Java SDK are different types even though their lexical names are the same. Thus, we replace the names with their type signatures (including their package and class names) for differentiation.
Similarly, function names are replaced by their full signatures.
For variable names, if they are non-primitive types, they are replaced by the type signatures, similar to class names; if they are primitive types, they are replaced by a type-specific token.
Tokens having no effect on code operational semantics, such as `{', `,', `;', are removed.
The below illustrates how three main kinds of tokens are normalized:

\begin{lstlisting}[basicstyle=\footnotesize,belowskip=0.5em,aboveskip=0.5em]
int i; ==> int int_id                             // 2 tokens
CommonTree ==> Antlr.Runtime.Tree.CommonTree      // 1 token
lexer.Emit();  ==> Antlr.Runtime.SlimLexer.Emit() // 1 token
\end{lstlisting}

\textbf{Add structural keywords for the tokens:} This is to add relevant AST node types for the tokens into the token streams so that the later learning steps may utilize more information implicit in raw code texts.
The below snippet illustrates this step:

\begin{lstlisting}[basicstyle=\scriptsize,belowskip=0em,aboveskip=0.5em]
Console.WriteLine("out"); ==> expr_stmt expr func_call System.Console.WriteLine(String) argument literal_type string  // 7 tokens
\end{lstlisting}

\subsection{The Bilingual skip-gram Model}

Our goal here is to learn distributed vector representations for cross-language code tokens, which can then serve as the basis for more complex composed code elements.
We use the bilingual skip gram model (BiSkip) \cite{luong2015bilingual} to achieve the goal.
The motivation behind BiSkip is to learn shared embeddings between tokens cross-lingually rather than just monolingually:
Rather than just predicting the tokens in one language, they use the tokens in one language to predict their aligned tokens in another language and vice versa.
For example, from a large corpus of Java and C\# code, the BiSkip model may be able to learn that the token \texttt{readonly} in C\# is aligned to and has the same meaning as the token \texttt{final} in Java, and {\tt final} often occurs together with {\tt public} and {\tt int} to define certain constants. Then, when given the token {\tt readonly}, we can use the BiSkip model to substitute {\tt final} for {\tt readonly} and predict that its surrounding tokens are \texttt{int} and \texttt{public}.
The BiSkip model has been shown to perform well for both bilingual and monolingual tasks \cite{luong2015bilingual}. In this paper, we utilize the Berkeley aligner \cite{liang2006alignment} to generate token alignments from the code token streams to be used by BiSkip.

\vspace*{-1ex}
\subsection{Hierarchical Models}
\label{approach:hierachical}
Once we get the vector representation for tokens across languages, we want to generate representations for more complex code elements, such as expressions, definitions, declarations, statements, methods, classes, and modules (Figure~\ref{figure:softwarelayers}) so that code mappings and program translations can be done for more complex elements.
Since all of the elements are hierarchical compositions of elements at lower levels of granularity including tokens, our intuition is to generate representations for elements at higher levels of granularity by composing the shared embeddings of their constituent elements.

According to \cite{kenter2015short}, simply averaging word embeddings of all words in a text can be a strong baseline for representing the whole text for the task of short text similarity comparison. Variants of this simple averaging strategy exist, such as averaging the embeddings with their weights measured in terms of term-frequency/inverse-document-frequency (TF-IDF) to decrease the influence of the most common words.
As a preliminary exploration in this paper, we only consider 3 levels of granularity of this task: expressions, statements, and methods, and use the simple averaging operation to compose shared embeddings according to the structures of code abstract syntax trees.
While the next section shows the limited settings produce promising code mapping results, we leave more comprehensive exploration of composition strategies for future work.

%% file: expr.tex
\subsection{Data}

Table \ref{tab:dataset} is a summary of our training dataset. We consider the language pair C\# and Java in all the evaluations. We collect the comparable dataset as in StaMiner \cite{DBLP:conf/kbse/NguyenNNN14}. We use the implementation of BiSkip from \cite{MultiVecLREC2016} to generate the vector representations of tokens.

\begin{table}
	\centering
{\footnotesize
		\begin{tabular}{c|ccc|ccc}
			\hline
			Project & \multicolumn{3}{c}{Java} & \multicolumn{3}{c}{C\#} \\
			\hline
			& Ver & Files & Methods & Ver & Files & Methods \\
			\hline
			Antlr(AN) & 4.0.0 & 276 & 3560 & 4.0.0 & 630 &  5049  \\
			db4o(DB) & 8.0 & 5556 & 38525 & 8.0 & 3845 & 23248 \\ 		
			fpml(FP) & 1.7 & 130 & 727 & 1.7 & 135 & 1038 \\ 	
			Itext(IT) & 7.0.5 & 1147 & 10003 & 7.0.5 & 2647 & 18842 \\ 	
			JGit(JG) & 4.10.0 & 1394 & 13862 & 4.10.0 & 1079 & 9203 \\
			JTS(JT) & 4.0.0 & 958 & 7883 & 4.0.0 & 1035 & 6640 \\
			Lucene(LC) & 7.1.0 & 6098 & 48038 & 7.1.0 & 2930 & 19961 \\
			POI & 4.0.0 & 3295 & 29172 & 4.0.0 & 2794 & 16717 \\	
			Neodatis(ND) & 2.1 & 960 & 10525 & 2.1 & 987 & 12153 \\		
            \hline
	\end{tabular}
	\caption{Overview of our training data set}
	\label{tab:dataset}
}
\end{table}

\subsection{Evaluation Tasks}
\subsubsection{Element mappings}
As described in Section \ref{approach:hierachical}, we aim to build vector representation for compositional cases in order to find good mappings in a hierarchical model across languages. We extract all the expressions, the statements and the methods of each project in our data set by traversing the AST representation to identify the element type, then we manually defined ground truth elements pairs. We treat each element in Java as a query to retrieve the top-k elements of C\#. Then we use Mean Average Precision (MAP) as the metric to evaluate this task. Due to the limitation of pages, instead of showing the MAP score for each type of element of each project, we calculate the {\em average MAP} of all project for each type of element. Table \ref{tab:levels_mapping} shows the results of the evaluations, with k = 1, 5, 10, respectively.

\begin{table}[t]
\centering
{\footnotesize
	\begin{tabular}{c||c|c|c}
		\hline
		Levels & Expressions & Statements & Methods \\
		\hline\hline
		Avg. MAP, k = 1 & 0.31 & 0.38 & 0.44 \\
		\hline
        Avg. MAP, k = 5 & 0.43 & 0.50 & 0.58 \\
		\hline
        Avg. MAP, k = 10 & 0.57 & 0.53 & 0.59 \\
		\hline
	\end{tabular}
	\caption{Average MAP scores for element mappings at various granularity levels}
	\label{tab:levels_mapping}
}
\end{table}

\subsubsection{API mappings}
We use the task described in StaMiner \cite{DBLP:conf/kbse/NguyenNNN14} to evaluate how effective our approach.
We consider 2 types of API names: classes and methods. For each \textit{name} in Java, we get its vector representation and use it as a query. The query is used to find the top-k nearest neighbors among the shared embeddings for C\# names. In this task, we only consider k = 1, which means we consider only the exact match of the query. Since there are too many APIs to build ground truth manually, we randomly select 100 APIs of each project for this task.
Table \ref{tab:precision_apis} shows the precisions for class mappings and method mappings.

Compared to StaMiner~\cite{DBLP:conf/kbse/NguyenNNN14} and DeepAM~\cite{Gu2017} that mine API mappings by using statistical machine translation techniques and deep learning, our work is more generalized in term of the kind of code elements supported. Their work only focuses on learning the mappings between language SDK APIs, while our approach allows mappings among any kind of structural code elements in a language beyond SDK APIs. 
Although finding the mappings for SDK APIs in different languages is a commonly needed and important task, developers 
often need more mappings for program elements (e.g., variable names, data structures, statements, method implementations, etc.).
We believe that being able to find the mappings among program elements of any granularity is a important step to reach the goal of automated language migration.

\begin{table}[t]
\centering
{\footnotesize
	\begin{tabular}{c||c|c|c|c|c|c|c|c|c}
		\hline
		Level$\backslash$Project & AN & FP & IT & JG & JTS & LC & POI & DB & ND\\
		\hline\hline
		Class & 0.85  & 0.82   & 0.88    & 0.69    &  0.83   &  0.82  & 0.78 & 0.86 & 0.79  \\
		\hline
		Method & 0.81   & 0.80    & 0.83   & 0.77    & 0.72    &   0.87 & 0.89 & 0.82  & 0.83  \\
        \hline
	\end{tabular}
	\caption{Precision of API method mappings}
	\label{tab:precision_apis}
}
\end{table}

We also found that our approach detects correctly about 400 more SDK API method mappings and 150 more SDK API class mappings that were not set in the latest mapping files in the Java2CSharp tool, while StaMiner detects 120 more SDK mappings for both classes and methods and 84 of which are also in our mappings.
In this evaluation, we only consider k=1 for the top-k nearest neighbors. We expect to find even more mappings if we consider a larger \textit{k} and we can find more mappings for APIs that are {\em not} in the language SDK libraries. We leave these evaluations for future work.

%% file: discussion.tex
\vspace*{-1ex}
\subsection{Threats to Validity}
For the model training, we use the same settings as described by Mikolov et al.~\cite{mikolov2013distributed}, which may not be the best with respect to our dataset. We will do more empirical research
to choose better hyper-parameters to improve training results.
The normalization step is mostly based on srcML \cite{collard2011lightweight} to get the AST of source code. At this moment, srcML supports four languages (\texttt{C, C++, C\#}, \texttt{Java}), and we only perform experiments on the Java -- C\# pair. In the future, we want to explore the generalizability of our approach with more programming languages supported by srcML and beyond.

The correctness of our API mappings results were checked by ourselves manually, which may be biased and incomprehensive. To evaluate the actual correctness and usefulness of our API mappings, we plan to do more large-scale evaluations by using Java2CSharp to see how our mappings can help reduce the compilation error rates when compared with StaMiner during actual migration of projects.

As described in \cite{DBLP:conf/sigsoft/TuSD14}, source code is very localized. Since we treat a whole file as a corpus, which means we ignore the localness of the tokens. The skip-gram model focuses on capturing the global regularities over the whole corpus, and neglects local regularities, thus ignoring the localness of software. A natural way to collect local-awareness parallel corpus is that we can slice the file into multiple slices based on dependence information, then we align the parallel corpus based on the similarity of the slices. 
In addition, we only aligned files based on filename similarity. We could do better based on file-content similarity.
Further, we constructed vector representations for higher level elements
by a simple averaging technique for all the tokens in such 
elements. We could do better based on 
more structural and semantic contexts of the tokens.
We leave more detailed explorations for future work.

%% file: conclusion.tex
This work proposes an approach to learn code element mappings across programming languages based on distributed vector representations. By utilizing the alignment of files in projects with multiple versions of implementations in different languages, we learn the alignment of tokens in an unsupervised manner, and generate the shared embeddings for tokens across languages.
Then, the shared embeddings for more complex code elements are composed from the embeddings of their constituent elements and tokens from bottom up according to the hierarchical structures of the syntax trees of the code.
Our preliminary evaluations show that our approach can map many code elements between Java and C\# accurately. This can serve as a foundation for more complicated tasks, such as program translation,
cross-language program classification, code clone detection, code reuse, and even synthesis.

In the future, we want to improve the techniques used in our approach and examine our approach for more language pairs.

%% file: ack.tex
This work is partly funded by a MOE Tier 1 Grant 17-C220-SMU-008 from SIS at SMU. We also thank the anonymous reviewers for 
their insightful comments and suggestions.